\documentclass[11pt]{article}

\usepackage[margin=1in]{geometry}
\usepackage[numbers]{natbib}
\usepackage{amsmath}
\usepackage{amssymb}
\usepackage{bm}
\usepackage[nameinlink]{cleveref}
\usepackage{graphicx}
\usepackage{subcaption}
\usepackage{url}

\title{Bayesian Complete-Pooling in Cross-Subject Classification for Motor Imagery Electroencephalogram}

\author{Ethan Davis \\ University of Washington, USA \\ \texttt{ethandavisecd@gmail.com} \\ ORCID: 0009-0005-7079-0858}

\date{}

\begin{document}

\maketitle

\begin{abstract}
Brain-computer interfaces (BCIs) have long sought calibration-free operation, yet classifiers are typically benchmarked on discrimination alone. Discrimination is blind to calibration, a meaningful gap given that electroencephalogram (EEG) signals are nonstationary and point-estimate classifiers can become overconfident under distribution shift. We conducted a large-scale study contrasting Bayesian complete-pooling models against frequentist baselines for cross-subject, left-hand versus right-hand motor imagery EEG classification across 20 datasets. Each of six frequentist pipelines was paired with an analogous Bayesian pipeline sharing identical, deterministic feature extraction, so that only the classifier was Bayesian. Bayesian classifiers were fit via Markov chain Monte Carlo posterior sampling. Our primary metric was the Brier score, decomposed into reliability and resolution, alongside the area under the receiver operating characteristic curve for discrimination and Shannon entropy for sharpness. For each metric we fit a random-effects meta-analysis with Knapp-Hartung adjustment, verified by leave-one-out influence analysis. Bayesian complete-pooling produced statistically significant improvements in reliability and increases in predictive uncertainty (lower sharpness), with 95\% confidence intervals of $[-0.0029, -0.0001]$ for reliability and $[0.0022, 0.0220]$ bits for entropy. Still, neither effect held statistical significance when datasets with flagged posterior sampling were excluded. Brier score, resolution, and discrimination showed no significant differences. Between-study heterogeneity was low across all metrics. Bayesian pipelines consumed roughly thirteen times more energy than their frequentist counterparts, a cost that remains modest relative to common household appliances. This modest cost makes partial-pooling across subjects and sessions, the structure complete-pooling omits, a feasible next step.
\end{abstract}


\section{Introduction}

A long-standing goal of brain-computer interface (BCI) research has been to develop calibration-free devices. Whether due to fatigue, hunger, or a drying out of sensor gel, the signal that comes out of current noninvasive brain measurement techniques changes with time. Worse, while human brains share a generic pattern of functionality, individuals vary significantly. The oldest, most reliable strategy to deal with session-to-session variation is training. Marginally better than months of training, we can simply recalibrate the BCI for each new usage. While somewhat tiresome for the subject, this can require as little as 10 minutes.

Small-scale studies based on maximum a posteriori (MAP) estimation for EEG classification have shown that partial-pooling across subjects and sessions is a promising direction of research \cite{Jayaram2018b}. Even for within-session BCIs, fully Bayesian multilevel/hierarchical modes have been shown to improve discrimination \cite{Zhang2018}. Complementary to these results, large-scale benchmarks of established (no-pooling and complete-pooling) BCI models based on maximum likelihood estimation (MLE) have uncovered ordered rankings of classifiers by name and by category in terms of discrimination \cite{Chevallier2024}.

Towards a large-scale experiment of the effects of partial-pooling on BCI prediction performance, we first estimated the effect of Bayesian complete-pooling for left hand and right hand motor imagery electroencephalogram (MI-EEG) classification contrasted with frequentist baselines. Complete-pooling is the non-hierarchical special case of a multilevel model with zero between-group variance \cite{Gelman2006}. Because it omits subject and session structure, it isolates the effect of posterior averaging from that of group-level structure. Any benefit of partial-pooling must therefore be measured against complete-pooling as its baseline. Each Bayesian pipeline also kept its control's deterministic feature extraction, isolating Bayesian classification from feature learning.

Discrimination metrics such as area under the receiver operating characteristic (AUROC) curve are by far the most common way to rank BCIs. However, discrimination is independent of calibration \cite{Ovadia2019}, and predictions can become badly miscalibrated under the distribution shift of nonstationary EEG. Posterior averaging moderates overconfident predictions \cite{Bishop2006}, which can improve calibration under such shift. Because this moderation scales with the predictive uncertainty of each trial, it can also reorder predictions and change discrimination. The size of either effect for MI-EEG classification had not been measured at scale.

In addition to discrimination, we contrasted frequentist pipeline controls to Bayesian pipeline treatments in terms of calibration. We used a proper scoring rule as our primary metric of comparison. As secondary metrics we analyzed calibration/reliability, resolution, discrimination, and sharpness \cite{Wilks2019}. Our experiment was a large-scale study that pooled binary classification results across 20 left hand and right hand MI-EEG datasets. The summary is that with Bayesian complete-pooling we found statistically significant differences in reliability and sharpness, but not in Brier score, resolution, or discrimination.

This study makes three contributions. First, theory predicts the direction of the calibration effect but not its size, nor whether discrimination changes, and prior Bayesian BCI studies were small-scale and focused on discrimination \cite{Jayaram2018b, Zhang2018}. We estimate these effects with confidence intervals across 20 datasets. Second, complete-pooling is the baseline for partial-pooling, which adds subject and session structure to the same model. Measuring it first means a later partial-pooling study can show what that structure adds beyond posterior averaging alone. Third, BCI benchmarks rank classifiers by discrimination alone \cite{Chevallier2024}. We report Brier score, its calibration and resolution components, and sharpness for six established pipelines, and release per-fold predictions for reuse.

In \Cref{Section:Background} we briefly overview preliminary concepts of Bayesian ML pointing to key characteristics that may benefit transfer learning research for BCIs. Additionally, we delineate the metrics of our experiment and why they are relevant for building trustworthy ML. Since our benchmark analysis includes discussions of effect sizes, uncertainty, and between-study heterogeneity measurements in addition to $p$-values, we explain these too. \Cref{Section:Methods} outlines our experiment design before \Cref{Section:Results} revealing the results and \Cref{Section:Discussion} where we interpret the outcomes, state limitations, and describe future work. \Cref{Section:Conclusion} summarizes our findings.

\section{Background}\label{Section:Background}

\subsection{Model Building}

In supervised learning, a task $T$ learns a mapping $f$ from inputs $\bm{x}\in\mathcal{X}$ to outputs $y\in\mathcal{Y}$. The inputs $\bm{x}$ are often a fixed dimensional vector $\bm{x}\in\mathbb{R}^D$ where $D$ is the dimensionality of the vector. An experience $E$ is given in the form of a set of $N$ input-output pairs $\mathcal{D}=\{(\bm{x}_n,y_n)\}^N_{n=1}$, known as the training set, where $N$ is the sample size. In classification problems, the output space is a set of $C$ unordered and mutually exclusive labels known as classes $\mathcal{Y}=\{\mathcal{C}_1,\mathcal{C}_2,\ldots,\mathcal{C}_C\}$. The problem of training is often to find a set of parameters that solves the maximum likelihood estimation of the model as in \Cref{Equation:Freq} \cite{Murphy2022}.

\begin{equation}
p(y \,|\, \bm{x},\hat{\bm{\theta}}) \quad \text{such that} \quad \hat{\bm{\theta}} = \operatorname*{argmax}_{\bm{\theta}} \sum_{n=1}^{N} \log p(y_{n} \,|\, \bm{x}_{n},\bm{\theta}).
\label{Equation:Freq}
\end{equation}

Bayesian ML lets us model a probability distribution over weights allowing us to represent uncertainty in our prediction. For a training set $\mathcal{D}$, output $y$, input $\bm{x}$, and model weights $\bm{\theta}$, we have the Bayesian treatment of ML found in \Cref{Equation:Bayes} \cite{Bishop2006}. However, there is a major drawback with the Bayesian framework which involves the integration over the space of parameters \cite{Bishop2023}. This is familiar from introductory calculus: It is easier to take a derivative, e.g. gradient descent, than it is to find an integral. Given a limited compute budget and ample training data, it is often better to use point estimation \cite{Bishop2023}, though, typically Bayesian ML corresponds to better generalization \cite{Murphy2023}.

\begin{equation}
p(y \,|\, \bm{x}, \mathcal{D}) = \int p(y \,|\, \bm{x},\bm{\theta}) p(\bm{\theta} \,|\, \mathcal{D}) \mathrm{d}\bm{\theta} \quad \text{where} \quad p(\bm{\theta} \,|\, \mathcal{D}) \propto p(\bm{\theta})p(\mathcal{D} \,|\, \bm{\theta}).
\label{Equation:Bayes}
\end{equation}

Uncertainty in parameters is naturally and coherently modeled with Bayesian ML via the posterior probability distribution $p(\bm{\theta} \,|\, \mathcal{D})$. Furthermore, this uncertainty estimation generalizes to multilevel/hierarchical models over datasets with group-level structure \cite{McElreath2020}. Indeed, multilevel modeling is a frequent motivation for using Bayesian methods. For BCIs, it is intuitive to see how multilevel modeling can be applied across subjects and sessions. Given the population-level nonstationarity of EEG signals, Bayesian ML offers useful estimation capabilities through principled uncertainty quantification approaches.

\subsection{Prediction Verification}

When building predictors our goal is to maximize sharpness subject to reliability \cite{Gneiting2007}. This concept intuitively refers to the accuracy and precision of a bullseye diagram where predictions may be tight/scattered and centered/off-centered. Formally, let $\hat{y}_m$ be a prediction which can take on any of the $M$ values $\hat{y}_1, \hat{y}_2, \ldots, \hat{y}_M$, and let $y_n$ be the corresponding target which may be any of the $N$ values $y_1, y_2, \ldots, y_N$. Accuracy generally refers to the average correspondence between individual predictions and the events they predict. Reliability, resolution, discrimination, and sharpness are components of accuracy \cite{Wilks2019}.

Reliability, or calibration, corresponds to the average observation $y_p$ given a prediction $\hat{y}_r$ summarizing the conditional distribution $p(y_p \,|\, \hat{y}_r)$. Resolution is also concerned with $p(y_p \,|\, \hat{y}_r)$ and measures the difference in conditional averages of an observation $y_p$ for different predictions $\hat{y}_r$ and $\hat{y}_s$. Discrimination is the converse of resolution, in that it pertains to differences between the conditional averages of a prediction $\hat{y}_r$ conditional on different targets $y_p$ and $y_q$. Sharpness is an attribute of predictions alone without regard to their corresponding observations, characterizing the unconditional distribution of a prediction $p(\hat{y}_r)$ \cite{Wilks2019}.

Prediction verification is easiest to understand with nonprobabilistic predictions of discrete predictands, such as accuracy calculated as the proportion of true positives and true negatives over all predictions. Verification of probabilistic predictions is more nuanced. Since nonprobabilistic predictions contain no expression of uncertainty, it is clear whether an individual prediction is correct or not. The Brier score (BS) is a common score for probabilistic predictions that is essentially the mean squared error of the predictions. BS averages the squared differences between pairs of predictions and observations shown in \Cref{Equation:BS}, where $K$ is the number of pairs \cite{Wilks2019}.

\begin{equation}
\mathrm{BS} = \frac{1}{K} \sum_{k=1}^{K} (y_k - \hat{y}_k)^2.
\label{Equation:BS}
\end{equation}

The Brier score decomposes into reliability, resolution, and uncertainty \cite{Murphy1973}. Let $K_m$ be the number of pairs with prediction $\hat{y}_m$, let $\bar{y}_m$ be the average observation given $\hat{y}_m$, and let $\bar{y}$ be the overall average observation. \Cref{Equation:Murphy} then holds for predictions taking the $M$ values $\hat{y}_1, \ldots, \hat{y}_M$ \cite{Wilks2019}. Reliability is lower when predictions match their conditional observed frequencies, and resolution is higher when those frequencies differ from the overall average. Ideally a model would minimize reliability and maximize resolution. Uncertainty is inherent to the data and cannot be optimized by the model.

\begin{equation}
\mathrm{BS} = \underbrace{\frac{1}{K}\sum_{m=1}^{M} K_m(\hat{y}_m-\bar{y}_m)^2}_{\text{reliability}} - \underbrace{\frac{1}{K}\sum_{m=1}^{M} K_m(\bar{y}_m-\bar{y})^2}_{\text{resolution}} + \underbrace{\bar{y}(1-\bar{y})}_{\text{uncertainty}}.
\label{Equation:Murphy}
\end{equation}

Prediction verification has perhaps been most developed by the atmospheric sciences, which have taken a calibration-first approach to research \cite{Wilks2019}. On the other hand, machine learning has largely developed through a discrimination-first approach \cite{Duda2001}. In the last several years, though, ML researchers have increasingly built calibration-aware models \cite{Guo2017}. Bayesian ML generally offers better calibrated models because they average over parameter space rather than optimizing point estimates which may result in overconfident (overly sharp) estimates \cite{Murphy2023}. This has motivated the use of Bayesian methods in high-risk scenarios for the purpose of trustworthy ML.

\subsection{Evidence Synthesis}

The goal of meta-analyses is to quantitatively synthesize evidence from multiple studies \cite{Harrer2021}. Meta-analyses of effect sizes are generally preferred since they address the issue of interest (the magnitude of the effect) rather than the null hypothesis (statistical significance) \cite{Borenstein2009}. An effect size captures the direction and magnitude of the relationship between two entities. For some study $k$, we denote the true effect $\theta_k$, the observed effect $\hat{\theta}_k$, and the sampling error $\epsilon_k$, such that we have \Cref{Equation:MetaAnalysis}. When pooling results, meta-analyses weight studies by their standard error (SE), defined by the standard deviation of the sampling distribution \cite{Harrer2021}.

\begin{equation}
\hat{\theta}_k = \theta_k + \epsilon_k.
\label{Equation:MetaAnalysis}
\end{equation}

\Cref{Equation:MetaAnalysis} is known as the random-effects model which assumes that there is not only one true effect size (fixed-effect model) but a distribution of true effect sizes. The goal of the random-effects model is to estimate the mean of the distribution of true effects. It stipulates that there is a second source of error $\zeta_k$ introduced by the fact that the true effect size $\theta_k$ is only part of an over-arching distribution of true effect sizes with mean $\mu$ giving us \Cref{Equation:RandomEffects}. A crucial assumption of the random-effects model is that the size of $\zeta_k$ is independent of $k$. This is the exchangeability assumption of the random-effects model \cite{Harrer2021}.

\begin{equation}
\hat{\theta}_k = \mu + \zeta_k + \epsilon_k.
\label{Equation:RandomEffects}
\end{equation}

The challenge associated with random-effects models is that we have to take the error $\zeta_k$ into account. To do this, we have to estimate the variance $\tau^2$ of the distribution of true effect sizes. It is ongoing research to find which estimator of $\tau^2$ performs best for different kinds of data. The DerSimionian-Laird estimator is based on closed-form expressions, while restricted maximum likelihood finds the optimal value of $\tau^2$ through an iterative algorithm. While significance tests of the pooled effect usually assume a normal distribution known as Wald tests, the Knapp-Hartung method is based on a $t$-distribution and used when the number of studies is small \cite{Harrer2021}.

$\tau^2$ is used to estimate the extent to which true effect sizes vary within a meta-analysis known as between-study heterogeneity. Extreme heterogeneity can mean that the studies have nothing in common, and that it makes no sense to interpret the pooled effect. The difficulty of quantifying between-study heterogeneity is to identify how much of the variation can be attributed to the sampling error, and how much to true effect size differences. The $I^2$ statistic is the percentage of variability in the effect sizes that is not caused by sampling error. Prediction intervals (PIs) give us a range into which we expect the effects of future studies to fall based on present evidence \cite{Harrer2021}.

Between-study heterogeneity can be caused by one or more studies with extreme effect sizes. When conducting a meta-analysis, we want to know if the pooled effect estimate is robust, meaning that it does not depend heavily on one single study. Such studies are called influential cases. For a meta-analysis with $K$ studies, the leave one out (LOO) method recalculates the results of the meta-analysis $K$ times, each time leaving out one study. Influence diagnostics allow us to detect the studies which influence the overall estimate of our meta-analysis the most, and let us assess if this large influence distorts our pooled effect \cite{Harrer2021}.

\section{Methods}\label{Section:Methods}

\subsection{Datasets}

We used the Mother of all BCI Benchmarks (MOABB) v1.5 to access all left hand and right hand MI-EEG datasets used in our experiment \cite{Jayaram2018}. For the purpose of unbiased data selection, we used all 20 available datasets under the conditions that, first, a dataset had at least 8 subjects and, second, it was not a subset of another dataset. There were 10 datasets that did not meet these criteria. \Cref{Table:SampleCharacteristics} shows an overview of the datasets used for our evaluation. Certain datasets were collected for the purpose of recording high-quality MI-EEG signals. Others were acquired such that specific experimental interventions could be tested.

\begin{table}
\caption{Sample characteristics of the 20 datasets.}
\centering
\begin{tabular}{lcccccc}
\hline
\multicolumn{1}{c}{Dataset} & No. subj. & No. chan. & Trial len. (s) & Freq. (Hz) & No. sess. & No. runs \\
\hline
BNCI2014\_001 \cite{Tangermann2012} & 9 & 22 & 4.0 & 250 & 2 & 6 \\
BNCI2014\_004 \cite{Tangermann2012} & 9 & 3 & 4.5 & 250 & 5 & 1 \\
Brandl2020 \cite{Brandl2020} & 16 & 63 & 4.5 & 1000 & 1 & 7 \\
Chang2025 \cite{Chang2025} & 28 & 59 & 4.0 & 1000 & 6 & 1 \\
Cho2017 \cite{Cho2017} & 52 & 64 & 3.0 & 512 & 1 & 1 \\
Dreyer2023 \cite{Pillette2021, Benaroch2022} & 87 & 27 & 5.0 & 512 & 1 & 6 \\
Forenzo2023 \cite{Forenzo2024} & 25 & 64 & 4.0 & 1000 & 2 & 3 \\
GrosseWentrup2009 \cite{GrosseWentrup2009} & 10 & 128 & 7.0 & 500 & 1 & 1 \\
GuttmannFlury2025\_MI \cite{GuttmannFlury2025} & 31 & 64 & 4.0 & 1000 & 3 & 1 \\
HefmiIch2025 \cite{Shi2025} & 37 & 32 & 10.0 & 256 & 6 & 1 \\
Kumar2024 \cite{Kumar2024} & 18 & 22 & 5.0 & 512 & 6 & 4 \\
Lee2019\_MI \cite{Lee2019} & 54 & 62 & 4.0 & 1000 & 2 & 1 \\
Liu2024 \cite{Liu2022, Liu2024} & 50 & 29 & 4.0 & 500 & 1 & 1 \\
PhysionetMI \cite{Goldberger2000} & 109 & 64 & 3.0 & 160 & 1 & 3 \\
Schirrmeister2017 \cite{Schirrmeister2017} & 14 & 128 & 4.0 & 500 & 1 & 2 \\
Shin2017A \cite{Shin2017} & 29 & 30 & 10.0 & 200 & 3 & 1 \\
Stieger2021 \cite{Stieger2021} & 62 & 62 & 3.0 & 1000 & 6 & 1 \\
Weibo2014 \cite{Yi2014} & 10 & 60 & 4.0 & 200 & 1 & 1 \\
Yang2025 \cite{Yang2025} & 51 & 59 & 4.0 & 1000 & 3 & 1 \\
Zhou2020 \cite{Zhou2021} & 8 & 41 & 5.0 & 500 & 7 & 9 \\
\hline
\end{tabular}
\label{Table:SampleCharacteristics}
\end{table}

\subsection{Paradigm}

\begin{figure}
\centering
\begin{subfigure}[c]{0.6\textwidth}
\centering
\includegraphics[width=\textwidth]{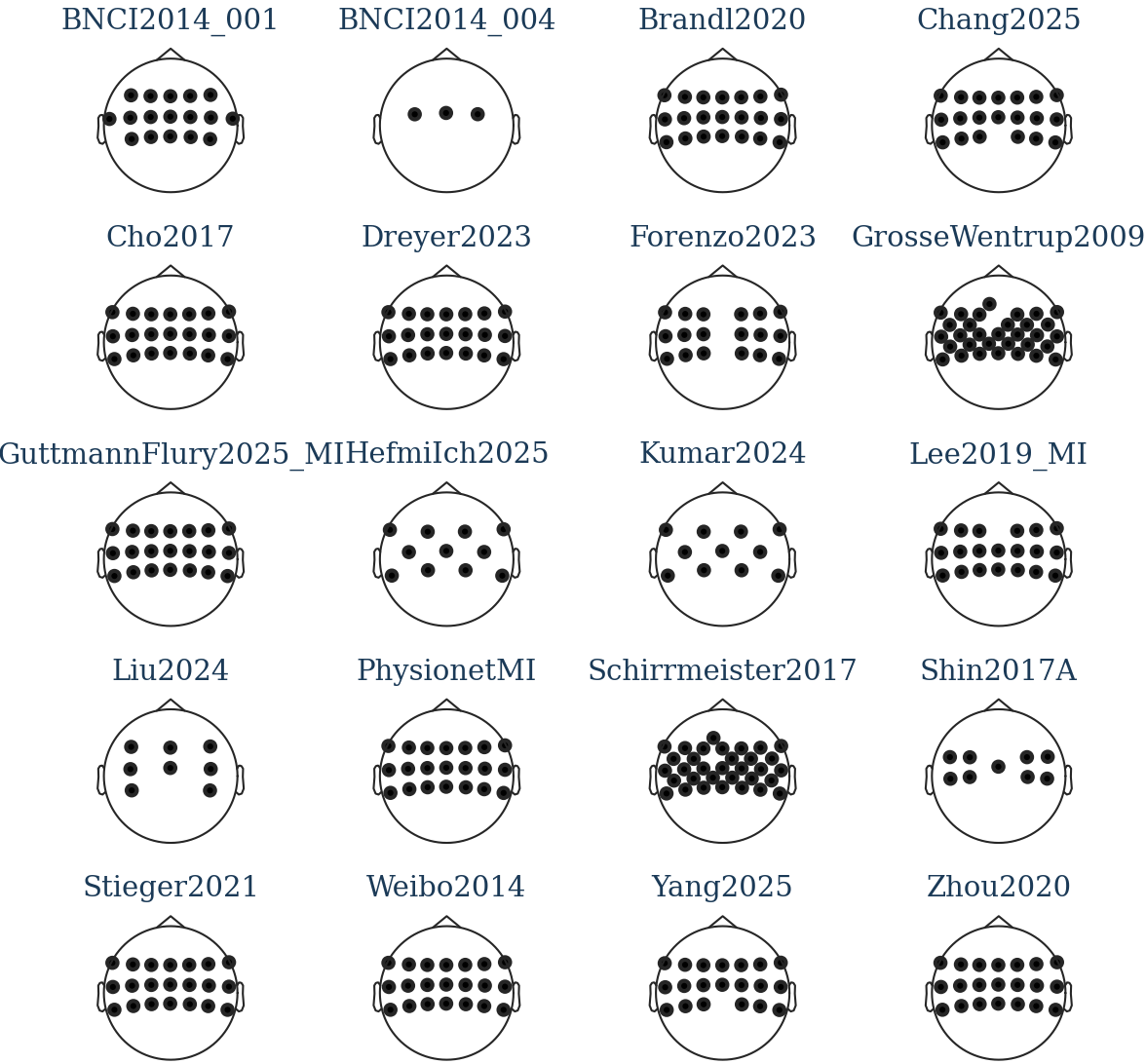}
\caption{Intersection of electrodes with the international 10-05 montage covering the sensorimotor cortex.}
\end{subfigure}
\vspace{1em}
\begin{subfigure}[c]{0.7\textwidth}
\centering
\includegraphics[width=\textwidth]{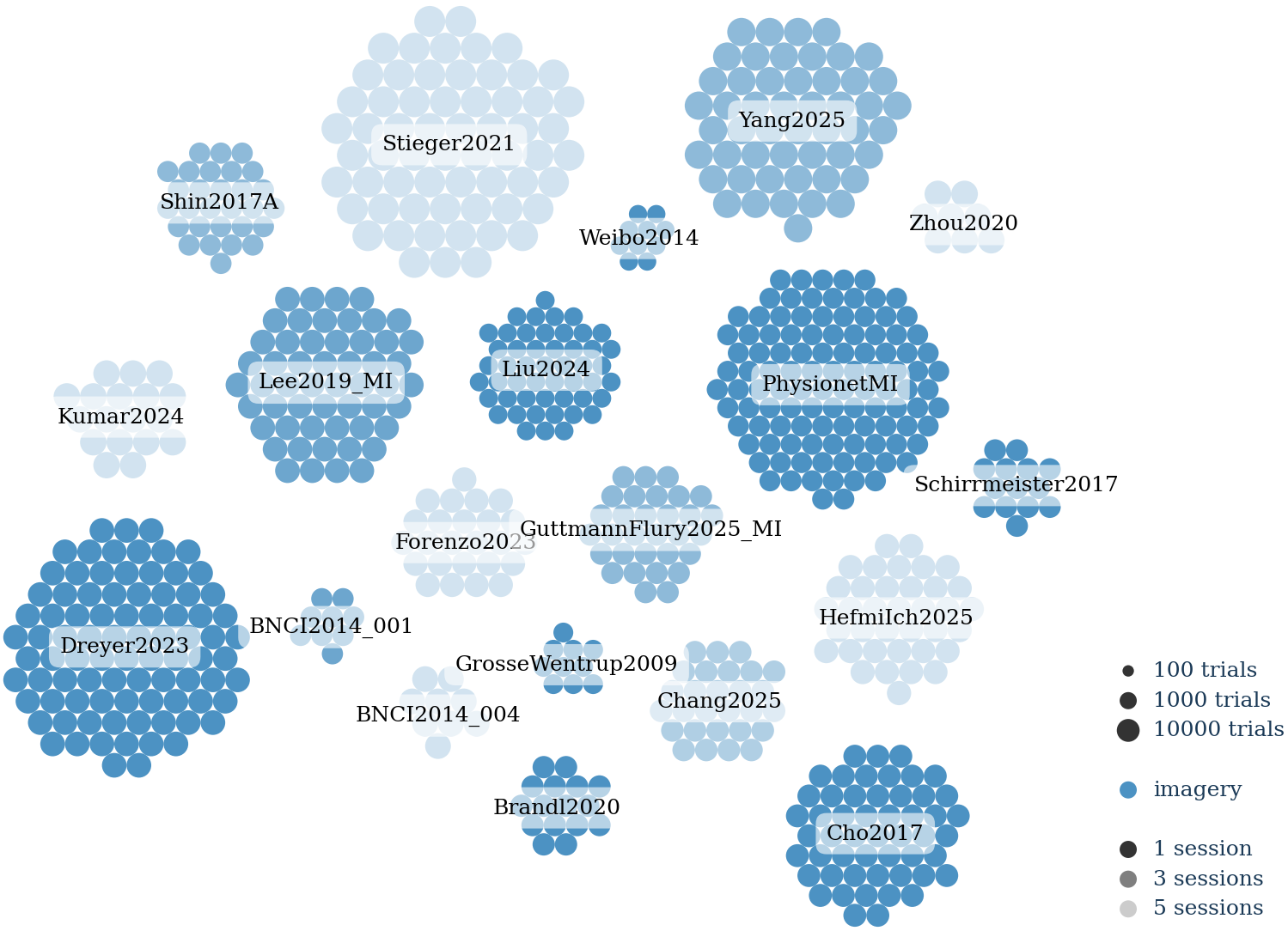}
\caption{Scale comparison in terms of number of sessions and total trials.}
\end{subfigure}
\caption{The MI-EEG datasets after preprocessing.}
\label{Figure:ParadigmsSpecs}
\end{figure}

Our experiment was concerned with left hand and right hand motor imagery and filtered any other classes from the datasets. These were the only classes shared by all 20 datasets and the classes on which \citet{Chevallier2024} ranked our control pipelines (\Cref{Section:Pipelines}). Additional filtering was the inclusion of only subjects 13--20 with Zhou2020, the A-subjects whose electrode information could be parsed. This exclusion was unrelated to computational resources. The eight remaining subjects still met our inclusion criterion in \Cref{Section:Methods} of at least 8 subjects per dataset. \Cref{Figure:ParadigmsSpecs} visualizes our datasets after signal preprocessing was completed.

For Stieger2021 we included only sessions 1--6 in response to GPU memory bottlenecks specifically in our Bayesian last-layer (BLL) networks. Every pipeline, Bayesian and frequentist, was evaluated on this same reduced dataset, so the exclusion could not favor either arm of the comparison. Doubling Stieger2021's retained trial count as an upper bound, the total trials in our experiment fell from at most approximately 221,900 to 178,478, at most a 19.57\% drop. Our leave-one-out influence analysis in \Cref{Section:Results} shows that omitting Stieger2021 entirely did not change the significance of any pooled effect.

Sensors were filtered for those belonging to the sensorimotor cortex of the international 10-05 montage. This subset of 33 channels spans the FC, FCC, C, CCP, and CP indexes. Filtering electrodes to this region reduced dimensionality in response to device memory exhaustion during training, while also reducing extraneous noise from other brain regions \cite{Nam2018}. Signals from these channels were then downsampled to 128 Hz, a common practice in BCI research \cite{Jayaram2018, Chevallier2024}. A 64 Hz Nyquist frequency still captures the 8--30 Hz mu/beta band where motor imagery signals live \cite{Corsi2023}.

\subsection{Pipelines}\label{Section:Pipelines}

Our work extends previous large-scale BCI benchmarking research by \citet{Chevallier2024} from within-session to cross-subject evaluation. In their study, three categories of BCI classifiers were identified: Raw, Riemannian, and Deep Learning. Respectively, these are pipelines based on raw signals, algorithms relying on Riemannian geometric statistics, and deep learning approaches. For their benchmark, 19 classifiers evaluated across 10 left/right hand MI-EEG datasets were used. Binary classification performance was ranked by the area under the receiver operating characteristic curve (AUROC) from within-session evaluation.

For our benchmark, we selected two of the top performing pipelines from each category of \citet{Chevallier2024}. From within both the Raw and Deep Learning categories we picked the first and second best performing pipelines out of nine and seven choices, respectively. With the Riemannian category, we chose the third and fourth highest performing pipelines out of seven options. We did not select either the first or second highest performing Riemannian pipelines because they did not map cleanly to a Bayesian ML treatment. \Cref{Table:FreqPipelines} shows the pipelines we selected that were drawn from \citet{Chevallier2024}.

\begin{table}
\caption{Baseline pipelines ordered by mean AUROC across 10 left/right hand MI-EEG datasets from \citet{Chevallier2024}, all used in our evaluation except Zhou2016 (4 subjects). Ranks among all 19 frequentist pipelines and AUROC are from within-session evaluation and were used only to select our control pipelines. Each control is shown with its paired Bayesian treatment. CSP: common spatial patterns; TS: tangent space; LDA: linear discriminant analysis; SVM: support vector machine; LR: logistic regression; SCNN/DCNN: ShallowConvNet/DeepConvNet; BLDA/BLR: Bayesian LDA/LR; GP: Gaussian process.}
\centering
\begin{tabular}{llccc}
\hline
\multicolumn{1}{c}{Control} & \multicolumn{1}{c}{Treatment} & Category & Orig. Rank & Avg. AUROC \\
\hline
TS+LR & TS+BLR & Riemannian & 3 & 81.479 \\
TS+SVM & TS+GP & Riemannian & 4 & 81.352 \\
SCNN & BSCNN & Deep Learning & 6 & 78.029 \\
CSP+SVM & CSP+GP & Raw Signals & 7 & 77.720 \\
CSP+LDA & CSP+BLDA & Raw Signals & 8 & 77.629 \\
DCNN & BDCNN & Deep Learning & 12 & 75.402 \\
\hline
\end{tabular}
\label{Table:FreqPipelines}
\end{table}

\subsubsection{Frequentist Baselines}\label{Section:FrequentistBaselines}

Both the common spatial patterns (CSP) and tangent space (TS) based classifiers depended on the covariance matrix of EEG channels across time points of each trial as input features. For covariance estimation, we used the oracle approximating shrinkage (OAS) estimator. Although its closed-form solution is sensitive to outliers, the low latency of OAS is ideal for real-time BCIs. With CSP we used the Euclidean metric for estimation of mean covariance matrices. Additionally, we decomposed 6 components along orthogonal discriminative axes. For TS we applied the Riemannian metric for reference matrix estimation and tangent space mapping.

For our linear discriminant analysis (LDA) classifiers we used the singular value decomposition (SVD) to fit Gaussian densities to left/right hand trials. This method avoids directly computing the covariance matrix which is numerically unstable. Our logistic regression (LR) configuration used the Limited-memory Broyden-Fletcher-Goldfarb-Shanno (L-BFGS) solver with default regularization. This is a second-order optimization algorithm, but uses limited memory to approximate the Hessian matrix. To accommodate convergence with complex datasets, we set the maximum number of iterations to 1000.

Support vector machines (SVMs) are generally not interpreted with probability calibration since their decision boundary is geometrically optimized. A common post-hoc method for measuring probabilistic metrics with SVMs is to use Platt scaling involving internal calibration cross-validation \cite{Deisenroth2020}. We used Platt scaling with 5-fold cross-validation when fitting our SVMs. Given CSP features, we trained SVMs with radial basis function (RBF) kernels. In contrast, for TS pipelines we applied a linear kernel. Geometrically, a linear decision boundary in the tangent space of the reference matrix approximates the best fit geodesic separating both classes.

To train ShallowConvNet (SCNN) and DeepConvNet (DCNN) \cite{Schirrmeister2017}, we used a uniform evaluation loop for both models. The loss function used was cross-entropy. The gradient update algorithm we used was AdamW. We configured the learning rate and weight decay to be $10^{-3}$ and $10^{-2}$, respectively. A maximum of 300 epochs with batch size of 64 and 80--20 validation split was used. Early stopping based on validation loss was applied with 150 epochs of patience and a $10^{-4}$ threshold for improvement. A learning rate scheduler was also used, with 50 epochs of patience before halving the learning rate, but not below $10^{-6}$, if the validation loss did not decrease.

\subsubsection{Bayesian Models}

The pipelines we selected from \citet{Chevallier2024} were the control group of our experiment. Each of these pipelines was paired with a Bayesian ML pipeline from our treatment group. In all pairs, feature engineering remained fixed. The pairwise contrast was frequentist point estimate optimization $p(y \,|\, x,\hat{\theta})$ versus Bayesian parameter space modeling of the posterior weights $p(\theta \,|\, \mathcal{D})$ with Bayesian model averaging for prediction $p(y \,|\, x, \mathcal{D}) = \int p(y \,|\, x, \theta) p(\theta \,|\, \mathcal{D}) \mathrm{d}\theta$. Indeed, all feature extraction was deterministic, whereas probabilistic classification was our experimental intervention.

We held feature extraction deterministic in every pipeline because Bayesian classification was the effect under test. A Bayesian treatment of features would have required a different model for each category: posteriors over spatial filters for CSP, over reference matrices on the manifold of covariance matrices for tangent space features, itself a separate direction of research in Riemannian statistics, and over every convolutional weight for SCNN and DCNN. Each would also have changed the features passed to the classifier, so a paired difference could no longer be attributed to classification. Fixed features instead gave every pair identical inputs and a single, common contrast.

Centering and scaling input features allowed our non-hierarchical prior specifications to generalize across datasets with heterogeneous feature scales. For all Bayesian pipelines, we applied $z$-score normalization to inputs after feature extraction and before classification. For the purpose of maintaining consistent feature engineering across all paired ML models, $z$-score transformation was also applied to features of frequentist pipelines. This experiment design localized the effect of Bayesian non-hierarchical prediction performance to classification.

We used Markov chain Monte Carlo (MCMC) based posterior sampling to fit our Bayesian pipelines. Specifically, No-U-Turn Sampling (NUTS) was applied. NUTS is an extension of Hamiltonian Monte Carlo (HMC) gradient-informed posterior sampling. For both warmup draws and tuning samples of MCMC we used 1000 iterations. To ensure robust sampling, we configured the target acceptance rate to be 0.95. The cost of a higher acceptance rate is often smaller steps per iteration and slower convergence, while the benefit is more accurate posterior approximations. All Bayesian pipelines were fit with 4 chains to validate the correctness of convergence.

\begin{figure}
\centering
\includegraphics[width=0.6\textwidth]{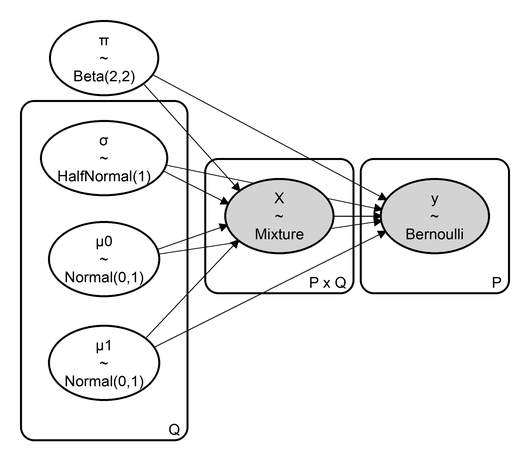}
\caption{The probabilistic graphical model of Bayesian linear discriminant analysis (BLDA), where $P$ is the number of trials and $Q$ is the number of electrodes.}
\label{Figure:BayesianPipelines}
\end{figure}

LDA is a type of Gaussian mixture model (GMM) that is supervised \cite{Hastie2009}. It was the only generative classifier in our experiment. To fit Bayesian LDA (BLDA), we modeled the likelihood of observations given labels $p(x \,|\, y)$ via a GMM and used Bayes theorem to estimate the probability of labels given observations $p(y \,|\, x)$. As a prior on the binary labels we used a weakly informative $\pi \sim \operatorname{Beta}(2,2)$ prior distribution. For the location priors and shared scale prior we used standard normal distributions $\bm{\mu}_0, \bm{\mu}_1 \sim \mathcal{N}(0,1)$ and a half-normal distribution $\bm{\sigma} \sim \mathcal{HN}(1)$, respectively. See \Cref{Figure:BayesianPipelines} for the probabilistic graphical model of our BLDA.

Bayesian logistic regression (BLR) appeared in one of our Riemannian pipelines, TS+BLR, and in both of our deep learning classifiers, BSCNN and BDCNN. We placed a standard normal prior on the bias, $b \sim \mathcal{N}(0,1)$, which was weakly informative on the logit scale. For the weights we relied on LeCun initialization such that $\sigma_w = \frac{1}{\sqrt{D}}$ where $D$ was the fan-in \cite{Murphy2023}, giving $\bm{w} \sim \mathcal{N}(\bm{0}, \sigma_w^2 \bm{I})$. Because $D$ was the number of features entering the classifier, it varied across pipelines and across datasets with different electrode sets. For $z$-scored inputs, this scaling kept the prior variance of the logits near one regardless of $D$.

In the deep learning pipelines, feature extraction was held fixed just as it was for CSP and tangent space features. In two stages we, first, trained the SCNN or DCNN architecture as in the frequentist pipeline and, second, dropped its deterministic linear classifier and fit a BLR predictor on the extracted features as a replacement. Each deep learning pipeline was therefore a BLL model that differed from its frequentist counterpart only in probabilistic classification \cite{Kristiadi2020, Brosse2020}. Last-layer Bayesian inference has been shown to mitigate the overconfidence of point-estimate networks \cite{Kristiadi2020}, while keeping MCMC tractable.

SVMs belong to a category of kernel methods \cite{Bishop2006}. As a Bayesian analog to SVMs we used Gaussian processes classifiers (GPCs). The MAP of GPCs with probit loss yields highly similar solutions to SVMs with its characteristic hinge loss function \cite{Rasmussen2005}. For CSP features with an RBF kernel, we used a half-normal amplitude prior $\eta \sim \mathcal{HN}(1)$ and a log-normal lengthscale prior $\ell \sim \mathcal{LN}(0,0.5)$. Given TS features and a linear kernel, we applied a half-normal amplitude prior distribution $\eta \sim \mathcal{HN}(1)$. To fit our GPCs we used the deterministic training conditional (DTC) method for sparse approximation with 100 inducing points found via $k$-means.

\subsection{Evaluation}

EEG signals have their lowest signal-to-noise ratio (SNR) when trials across sessions and subjects are pooled. We fit Bayesian non-hierarchical models for cross-subject classification to test the effect of our treatment in this scenario. Our use of cross-subject evaluation as opposed to within-session evaluation or cross-session (within-subject) evaluation provided an environment where the benefits of uncertainty quantification in pattern recognition from Bayesian ML would be most visible. Cross-subject classification is also the setting a calibration-free BCI must operate in, since it predicts for subjects absent from training.

Our experiment used a combination of leave one subject out (LOSO) and 10-fold cross-validation. We used LOSO if the number of subjects in a dataset was less than 10, and otherwise 10-fold CV grouped by subject (GroupKFold). In both schemes, each subject's trials appeared only in the training or only in the test partition of a fold, and any validation data was drawn from the training partition \cite{Jayaram2018}. Folds were fixed for each dataset, so every Bayesian pipeline and its frequentist control were tested on identical trials, making each pairwise difference a matched comparison.

Our primary metric for pairwise comparisons was Brier score. As a proper scoring rule, BS awards the true beliefs of a predictor, avoiding hedging, or gaming, of predictions \cite{Wilks2019}. BS can be decomposed into reliability and resolution \cite{Murphy1973}, and we used these components directly as secondary metrics to measure prediction performance. To decompose BS we used the CORP (Consistent, Optimally binned, and Reproducible) framework \cite{Dimitriadis2021}. We also measured discrimination and sharpness for prediction verification. For discrimination we measured AUROC, and for sharpness we measured Shannon entropy \cite{MacKay2003}.

\subsection{Analysis}

The 20 datasets we used had varying hardware, subjects, and samples meaning they were heterogeneous and could not be merged into one large dataset. For each dataset, we computed mean LOSO or 10-fold CV scores for all of our pipelines, before pairwise differences. Each Bayesian pipeline was paired with its frequentist counterpart (\Cref{Section:FrequentistBaselines}), giving six matched pairs per dataset. We computed this pairwise difference for each of our 5 metrics, generating 6 data points per dataset per metric, or 120 pairwise comparisons across all 20 datasets.

Effect sizes and SEs were computed for matched groups \cite{Borenstein2009}. For each of our 5 metrics, the six pairwise differences per dataset were combined into one effect size and standard error per dataset, giving the input to our meta-analysis. We fit a random-effects meta-analysis to these per-dataset effect sizes, using a restricted maximum likelihood (REML) estimator for the between-dataset variance and a Knapp-Hartung adjustment to the standard error of the pooled estimate. We also performed a leave-one-out influence analysis, refitting the model with each dataset removed in turn, to verify that our pooled estimates were not driven by any single dataset.

\section{Results}\label{Section:Results}

\subsection{Posterior Sampling}

\begin{table}
\caption{Convergence diagnostics from posterior sampling totaled across 196 runs for each pipeline.}
\centering
\begin{tabular}{lcccccc}
\hline
\multicolumn{1}{c}{Pipeline} & $\mathrm{ESS}_{\text{bulk}} < 400$ & Min. $\mathrm{ESS}_{\text{bulk}}$ & $\mathrm{ESS}_{\text{tail}} < 400$ & Min. $\mathrm{ESS}_{\text{tail}}$ & $\hat{R} > 1.01$ & Max. $\hat{R}$ \\
\hline
BDCNN & 0 & 3157 & 0 & 1777 & 0 & 1.01 \\
BSCNN & 0 & 3389 & 0 & 1442 & 0 & 1.01 \\
CSP+BLDA & 0 & 959 & 0 & 914 & 0 & 1.01 \\
CSP+GP & 6 & 272 & 1 & 359 & 1 & 1.02 \\
TS+BLR & 0 & 2371 & 0 & 1749 & 0 & 1.01 \\
TS+GP & 8 & 160 & 1 & 253 & 6 & 1.02 \\
\hline
\end{tabular}
\label{Table:PosteriorSampling}
\end{table}

For our Bayesian pipelines, we measured posterior sampling diagnostics to verify the quality of our estimates: bulk and tail effect sample size (ESS), and potential scale reduction factor ($\hat{R}$). ESS takes autocorrelation of our draws into account and provides the number of draws we would have if our sample was iid. $\hat{R}$ estimates convergence of separate chains estimating their similarity at the end of sampling. The recommended threshold for ESS bulk/tail is above 400, and values of $\hat{R}\leq1.01$ are considered safe \cite{Martin2021}. Grouping by pipelines, we aggregated worst case diagnostics across datasets and cross-validation in \Cref{Table:PosteriorSampling}.

Flagged runs were confined to our two Gaussian process pipelines and to 6 of the 20 datasets. For TS+GP these were Dreyer2023, HefmiIch2025, and Stieger2021, and for CSP+GP these were Forenzo2023, Liu2024, and Zhou2020. As a sensitivity check, we repeated each meta-analysis without these 6 datasets. Pooled estimates were essentially unchanged, at $-0.0016$ for reliability and 0.0131 bits for entropy, compared with $-0.0015$ and 0.0121 bits on all 20 datasets. Neither effect remained statistically significant, with confidence intervals of $[-0.0037,0.0005]$ and $[-0.0056,0.0317]$ bits, respectively.

\subsection{Meta-Analysis}

The pooled effect of accuracy (Brier score) was not significant. Additionally, both the pooled effect of resolution and discrimination (AUROC) were not significant. However, the aggregated effect size of reliability and sharpness were statistically significant. For reliability, the direction of the effect was negative meaning that Bayesian complete-pooling reduced calibration error. Furthermore, the direction for sharpness (Shannon entropy) was positive telling us that Bayesian complete-pooling increased uncertainty in predictions. The 95\% CIs were $[-0.0029, -0.0001]$ for reliability and $[0.0022, 0.0220]$ bits for entropy. See \Cref{Table:PooledEffects} for details.

\begin{table}
\caption{The pooled effects for all metrics.}
\centering
\begin{tabular}{lcccccc}
\hline
\multicolumn{1}{c}{Metric} & Estimate & SE & $t$(19) & $p$ & 95\% CI LB & 95\% CI UB \\
\hline
Brier Score & -0.0010 & 0.0007 & -1.3016 & 0.2086 & -0.0025 & 0.0006 \\
Reliability & -0.0015 & 0.0007 & -2.2100 & \textbf{0.0396} & -0.0029 & -0.0001 \\
Resolution & -0.0004 & 0.0003 & -1.5229 & 0.1442 & -0.0009 & 0.0001 \\
AUROC & -0.0002 & 0.0009 & -0.2586 & 0.7987 & -0.0021 & 0.0017 \\
Shannon Entropy & 0.0121 & 0.0047 & 2.5504 & \textbf{0.0195} & 0.0022 & 0.0220 \\
\hline
\end{tabular}
\label{Table:PooledEffects}
\end{table}

The variance components of between-study heterogeneity measured by our meta-analyses suggested that our results across left/right hand MI-EEG datasets were homogenous across datasets. That is, according to the rule of thumb for how heterogeneity is interpreted through $I^2$ (25\% = low, 50\% = moderate, 75\% = substantial) \cite{Higgins2002}. These outcomes validate the 95\% PIs that we computed. PIs are generally wider than the 95\% confidence intervals (CIs) of the pooled effect estimate. Statistically significant estimates with PIs that cross zero are less robust. This is what we see from reliability, though, the pooled effect of sharpness remains positive in \Cref{Table:Variance}.

\begin{table}
\caption{The between-study heterogeneity.}
\centering
\begin{tabular}{lcccccc}
\hline
\multicolumn{1}{c}{Metric} & $Q$(19) & $p$ & $I^2$ & $\tau$ & 95\% PI LB & 95\% PI UB \\
\hline
Brier Score & 23.4194 & 0.2194 & 22.74\% & 0.0014 & -0.0043 & 0.0024 \\
Reliability & 23.1407 & 0.2312 & 28.61\% & 0.0014 & -0.0047 & 0.0018 \\
Resolution & 17.4359 & 0.5604 & 3.05\% & 0.0002 & -0.0011 & 0.0003 \\
AUROC & 17.5546 & 0.5523 & 20.65\% & 0.0020 & -0.0049 & 0.0044 \\
Shannon Entropy & 18.7605 & 0.4723 & 0.45\% & 0.0015 & 0.0017 & 0.0225 \\
\hline
\end{tabular}
\label{Table:Variance}
\end{table}

Across all LOO iterations, Brier score, resolution, and AUROC showed null pooled differences and low between-study heterogeneity. However, for reliability, $11/20=55\%$ of iterations yielded null difference pooled effects, suggesting this result was sensitive. In the remaining 45\% of iterations the pooled effect was statistically significant, and in all cases heterogeneity remained low. For Shannon entropy, just 1/20 iterations yielded a null pooled effect, and every iteration had low heterogeneity.

\begin{figure}
\centering
\includegraphics[width=0.8\textwidth]{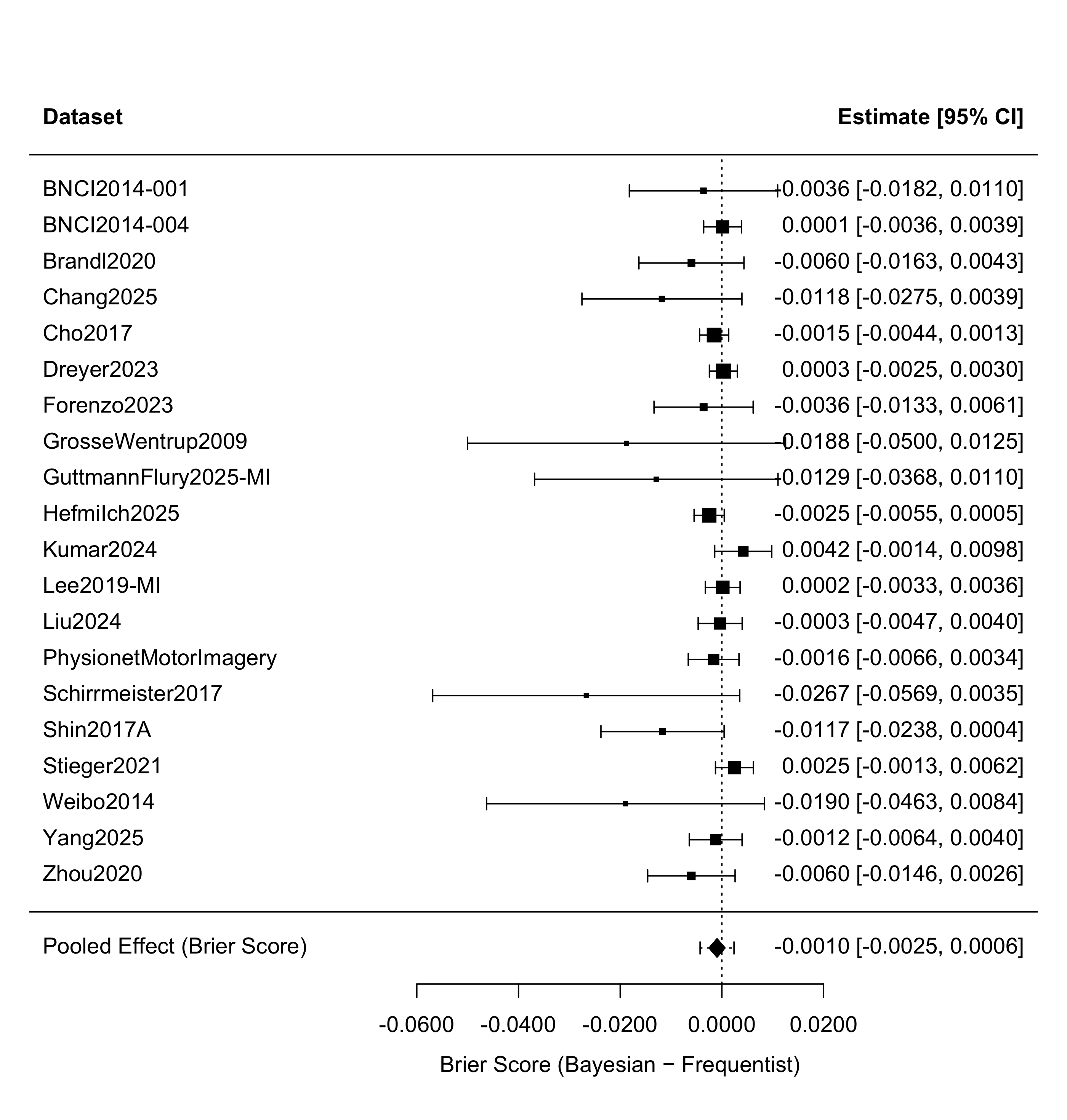}
\caption{The Brier score forest plot.}
\label{Figure:BrierForest}
\end{figure}

\begin{figure}
\centering
\begin{subfigure}[c]{0.8\textwidth}
\centering
\includegraphics[width=0.8\textwidth]{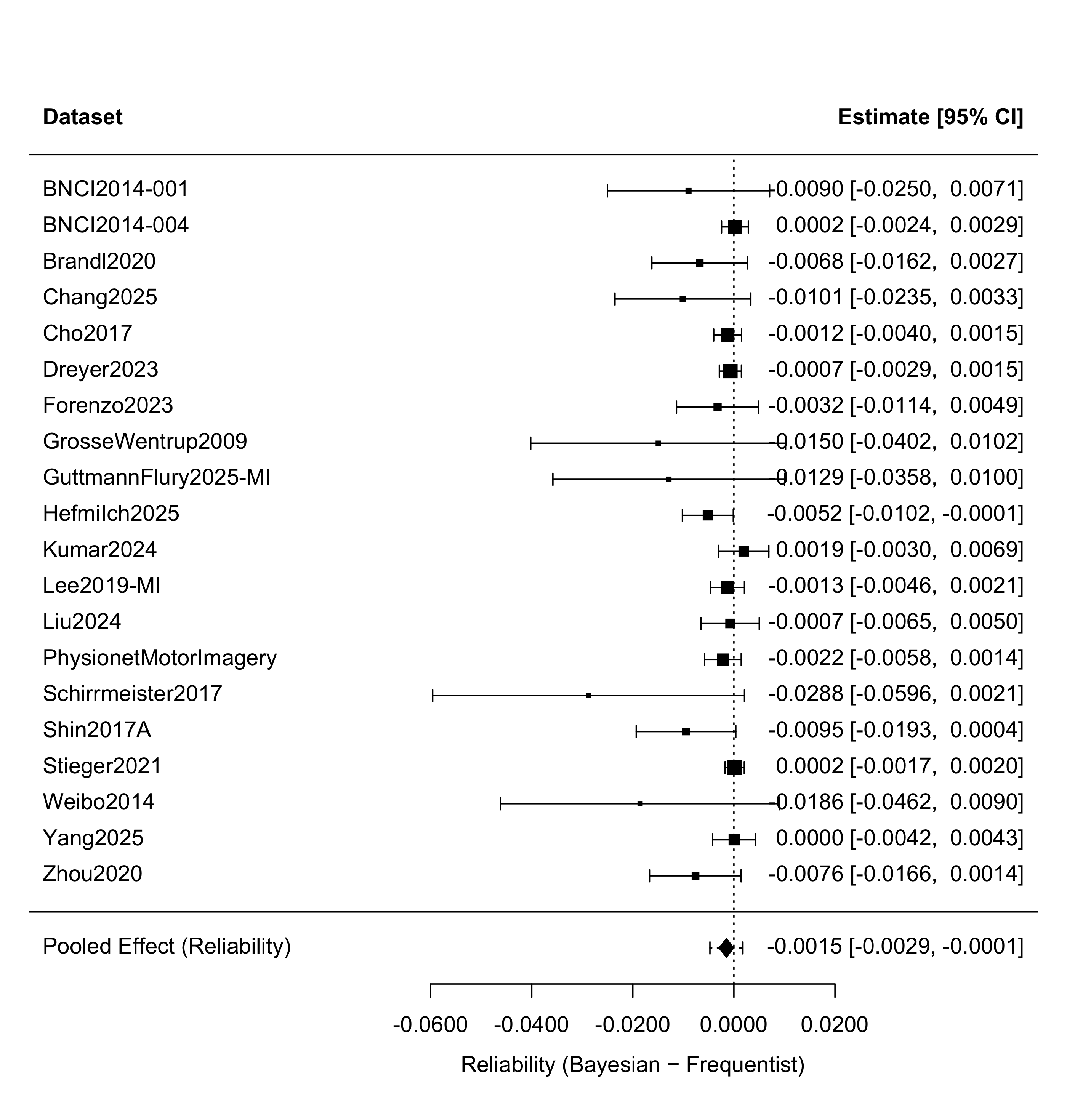}
\caption{Per-dataset effects, pooled estimate, and prediction interval.}
\end{subfigure}
\vspace{1em}
\begin{subfigure}[c]{0.8\textwidth}
\centering
\includegraphics[width=0.8\textwidth]{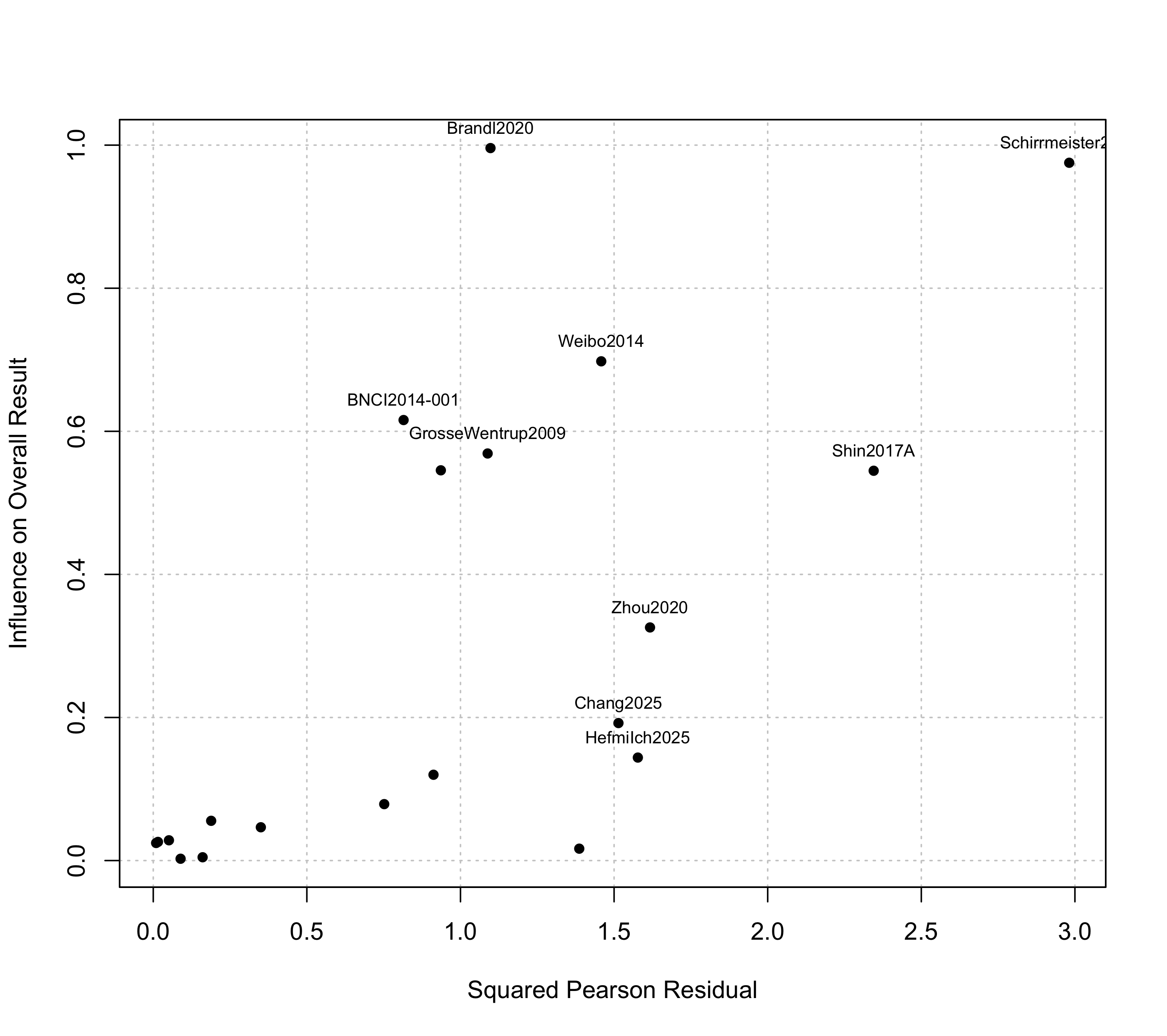}
\caption{Contribution to heterogeneity versus influence on the pooled result.}
\end{subfigure}
\caption{Forest and Baujat plots for reliability.}
\label{Figure:Reliability}
\end{figure}

\begin{figure}
\centering
\begin{subfigure}[c]{0.8\textwidth}
\centering
\includegraphics[width=0.8\textwidth]{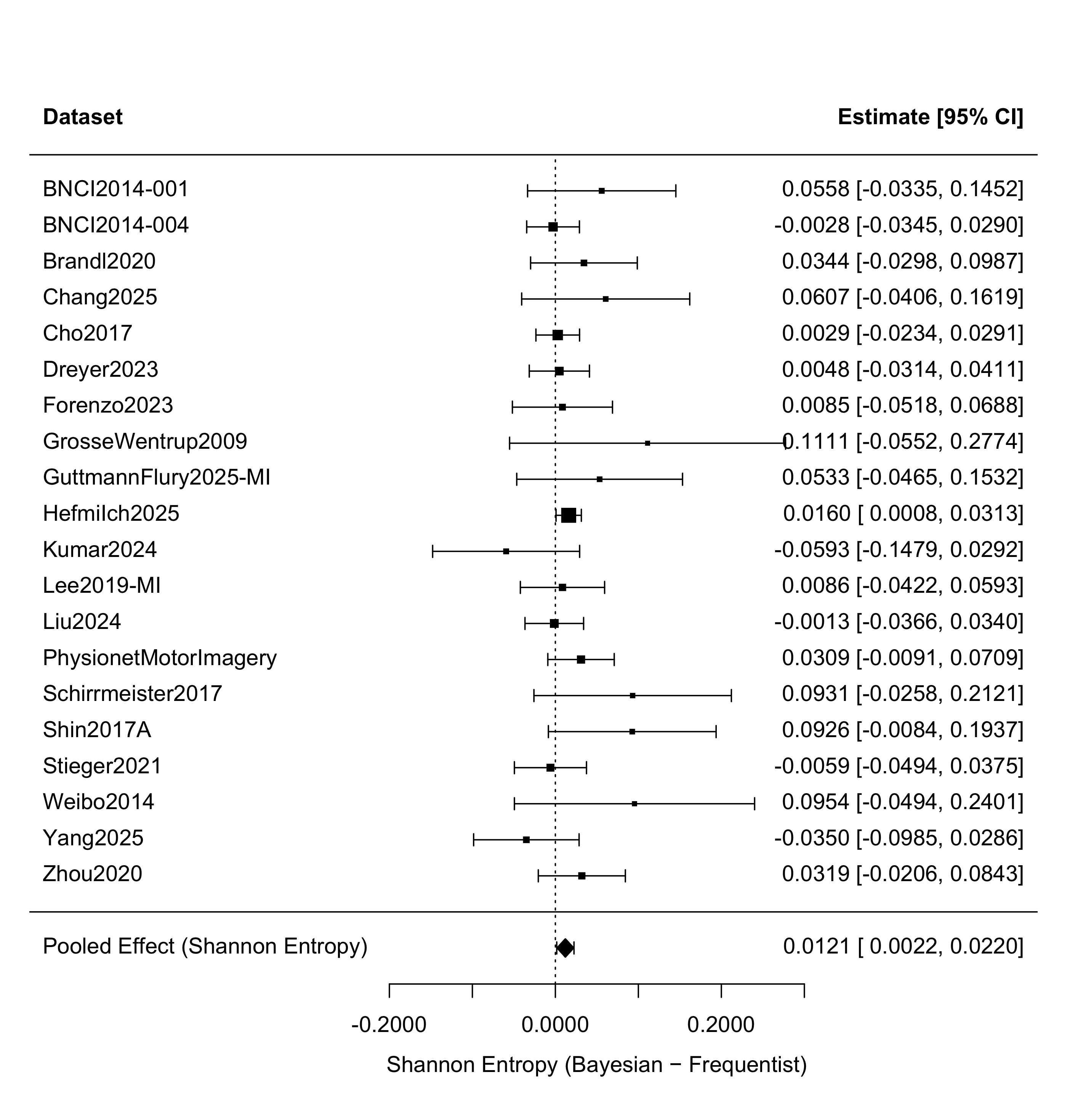}
\caption{Per-dataset effects, pooled estimate, and prediction interval.}
\end{subfigure}
\vspace{1em}
\begin{subfigure}[c]{0.8\textwidth}
\centering
\includegraphics[width=0.8\textwidth]{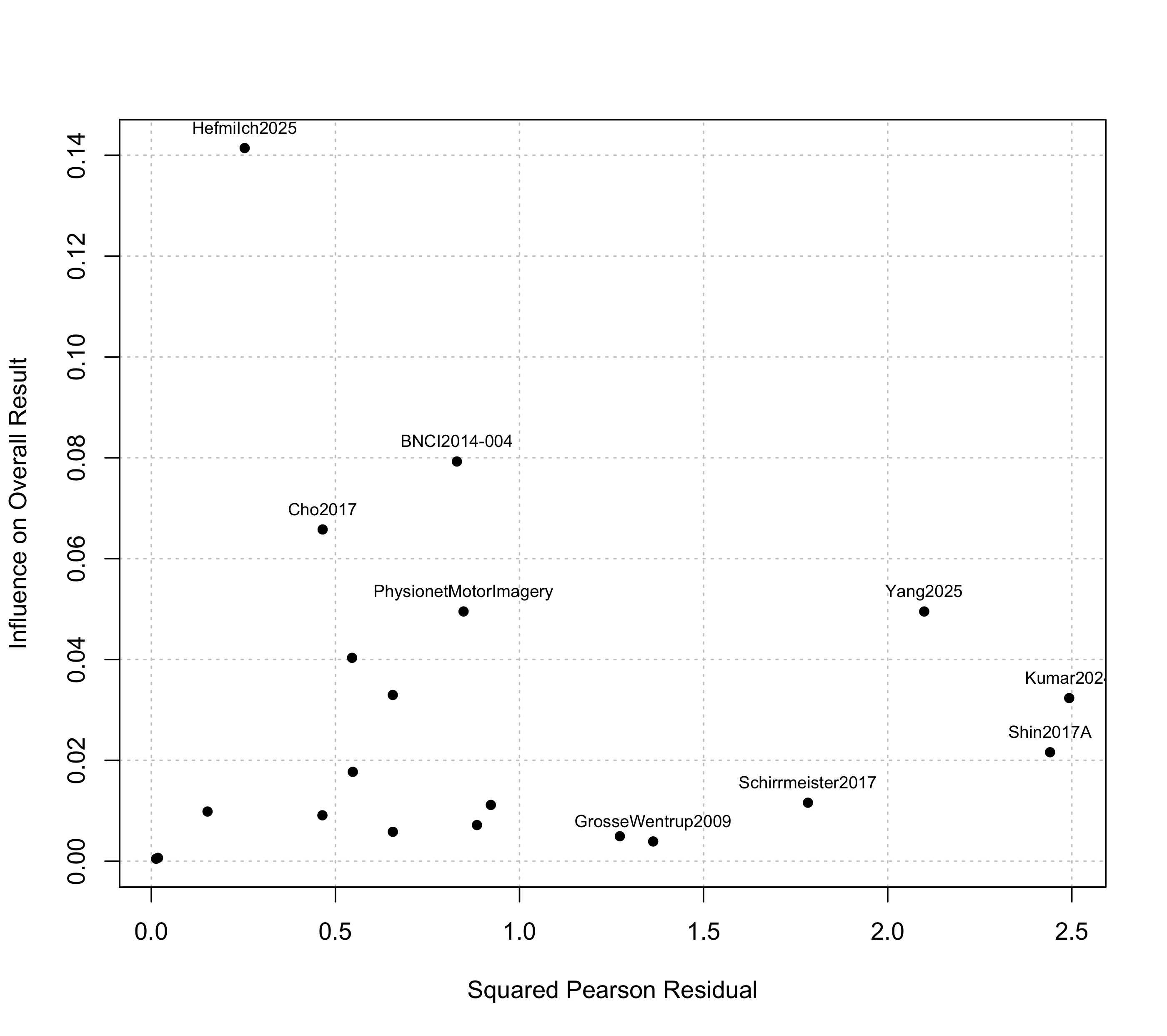}
\caption{Contribution to heterogeneity versus influence on the pooled result.}
\end{subfigure}
\caption{Forest and Baujat plots for Shannon entropy.}
\label{Figure:Entropy}
\end{figure}

\Cref{Figure:BrierForest} shows the forest plot from our meta-analysis of Brier score. Additionally, \Cref{Figure:Reliability} and \Cref{Figure:Entropy} show the forest plots for reliability and Shannon entropy, respectively. For each study in a forest plot, the size of the square corresponds to the precision of its point estimate, driving the weight of the study when estimating the pooled effect. The brackets represent the 95\% confidence intervals of a study. Likewise, the size of the diamond corresponds to the precision of the pooled effect estimate. However, its brackets represent the 95\% prediction intervals, denoting where we would expect to find the effect size of an unseen study \cite{Harrer2021}.

\Cref{Figure:Reliability} and \Cref{Figure:Entropy} also show Baujat plots from our influence analysis of reliability and Shannon entropy, respectively. In a Baujat plot, the horizontal axis corresponds to the contribution of a study to the overall between-study heterogeneity, and the vertical axis pertains to how much the overall heterogeneity would change if the study were removed. A point in the upper-right is a study that is both a poor fit to the model and substantially shifts the pooled estimate when removed, the classic signature of a problem study the plot is designed to surface \cite{Harrer2021}. In our Baujat plots, extreme points in the top quartile on either axis are labeled.

\subsection{Compute Profiling}

\begin{table}
\caption{Central tendencies of energy consumption (Wh) across datasets. Prior to central tendency calculation, mean energy consumption was calculated between folds of LOSO/CV within datasets.}
\centering
\begin{tabular}{lccc}
\hline
\multicolumn{1}{c}{Pipeline} & Q1 & Median & Q3 \\
\hline
CSP+LDA & 0.747 & 0.868 & 0.981 \\
CSP+BLDA & 30.877 & 51.137 & 76.378 \\
CSP+SVM & 0.822 & 1.084 & 1.874 \\
CSP+GP & 48.846 & 64.333 & 133.244 \\
TS+LR & 0.797 & 1.235 & 1.781 \\
TS+BLR & 8.525 & 12.924 & 29.590 \\
TS+SVM & 1.707 & 4.675 & 18.620 \\
TS+GP & 21.665 & 36.484 & 52.943 \\
SCNN & 6.782 & 13.125 & 23.696 \\
BSCNN & 40.347 & 59.164 & 91.962 \\
DCNN & 6.815 & 10.284 & 19.558 \\
BDCNN & 26.534 & 40.923 & 92.504 \\
\hline
\end{tabular}
\label{Table:Energy}
\end{table}

\Cref{Table:Energy} summarizes the energy consumption measured in watt-hours (Wh) from training pipelines across datasets. As context, consider that the French Agency of Ecological Transition reports that the cost of charging a smartphone once per day is 19.178 Wh/day \cite{ADEME2026}. Generally, our frequentist pipelines were below this marker, and our Bayesian pipelines were above it. To gauge the cost of charging a smartphone, consider that a combination refrigerator freezer with 24/7 usage on average consumes 863.01 Wh/day, and running a washing machine and tumble dryer together has an average combined energy consumption of 1338.38 Wh/cycle \cite{ADEME2026}.

Computed in log space, the geometric-mean ratio of increased energy consumption by the Bayesian complete-pooling pipelines was $13.1\times$. These verifications gave us perspectives to quantitatively understand the computational costs of training Bayesian complete-pooling models for MI-EEG classification, both in terms of comparisons to baseline frequentist models, and in absolute terms. Bayesian ML is substantially more expensive to train than baseline models, however, compared to everyday household appliances it is economical. Since training costs heavily depend on algorithm and hardware implementations, we kept our analysis of computational costs informal.

\section{Discussion}\label{Section:Discussion}

\subsection{Interpretation}

Our experiment found statistically significant effects of Bayesian complete-pooling on cross-subject MI-EEG classification only for sharpness and reliability. The clearer significant effect was sharpness: Shannon entropy increased, so Bayesian predictions were more hedged between classes. This effect held in 19 of 20 leave-one-out iterations, with low heterogeneity throughout (\Cref{Section:Results}). Without the 6 datasets with flagged posterior sampling, its pooled estimate held but was no longer statistically significant. We therefore treat sharpness as our clearest, though not conclusive, evidence that Bayesian classification does real work under complete-pooling.

Reliability also improved on average, with predicted class frequencies matching true frequencies more closely than under our frequentist baselines. This effect was far less robust: removing any one dataset from our leave-one-out analysis made it statistically null in 11 of 20 iterations (\Cref{Section:Results}). It also lost significance without the 6 datasets with flagged posterior sampling. We therefore treat reliability as a secondary, exploratory finding rather than firm evidence for a calibration benefit of complete-pooling alone.

The effects on sharpness that we empirically observed tell us that our baseline frequentist classifiers were overconfident. Bayesian model averaging integrates predictions across the full posterior rather than relying on a single point estimate. In high-dimensional parameter spaces, the region immediately around the posterior mode contains disproportionately little of the total probability mass compared to the much larger volume of the surrounding region, known as the typical set \cite{MacKay2003}. Therefore, the averaged predictive distribution tends to smooth out the overconfident predictions a point estimate alone would produce.

At the same time, the null differences we observed for resolution and discrimination tell us that our Bayesian complete-pooling models found a similar decision boundary to our frequentist baselines. Like our point estimate models, the Bayesian complete-pooling models averaged decision boundaries across the population of subjects and sessions for low SNR MI-EEG. This population-level averaging explains why uncertainty quantification did not change classification accuracy. Our null difference in Brier score tells us that Bayesian complete-pooling does not improve the ability of classifiers to recognize patterns in left vs. right hand MI-EEG.

Our null difference in discrimination is consistent with prior small-scale results on pooling. Partial-pooling across subjects and sessions has shown promise in MAP-based studies \cite{Jayaram2018b}, and fully Bayesian hierarchical models have improved within-session discrimination \cite{Zhang2018}. Both approaches add group-level structure, which our complete-pooling models omitted by design. Any discrimination benefit of Bayesian ML therefore likely depends on that structure rather than on posterior averaging alone. Cross-subject classification is the setting a calibration-free BCI must operate in. In this setting, posterior averaging alone did not improve discrimination.

\subsection{Limitations}

Our pipeline evaluations were fit without hyperparameter optimization (HPO). Hyperparameter values were instead set from the literature and from our own experiment design. Our Bayesian models placed weakly informative priors on the analogous parameters, inferring them jointly with the model via MCMC rather than fixing them to a single value. Grid search over 12 pipelines fit with LOSO or 10-fold cross-validation across 20 datasets, plus inner 5-fold Platt scaling for the SVMs, was also impractical under our computational budget \cite{Kuhn2019}. Libraries such as Optuna could make HPO feasible in future work \cite{Ozaki2025}.

The pipelines most exposed to this gap are our four neural networks, SCNN, DCNN, BSCNN, and BDCNN, whose many parameters and nonlinear loss landscapes are more sensitive to optimization, yet learning rate and weight decay were fixed at $0.001$ and $0.01$ across 20 heterogeneous datasets. Our Gaussian process pipelines, CSP+GP and TS+GP, are less exposed: their kernel hyperparameters were assigned priors and inferred via MCMC, though a weak prior can still hurt fit. The SVM regularization constant is a single low-sensitivity scalar, and our LDA and logistic regression pipelines have no comparable hyperparameter.

We tested Bayesian complete-pooling only and did not fit partial-pooling models. Complete-pooling served as a low-cost baseline for partial-pooling, which would have added group-level parameters and hyperpriors for every subject and session. These additional parameters would have enlarged the posterior, making sampling slower and more costly \cite{McElreath2020}. Because partial-pooling models the subject and session structure that complete-pooling omits, it is the natural next step. Whether partial-pooling yields larger effects on prediction performance, as we hypothesize, is left to future experiments.

Meta-analysis was developed for the purpose of accumulating results from multiple studies where raw data is not available for fitting regression models. Summary statistics reported by studies, though, are lossy summaries of the information available in raw data. Indeed, our analysis of prediction performance was not limited to a two-stage approach that calculated effect sizes before estimating pooled effects. Instead, a one-stage approach to multilevel random-effects models could have been taken that was directly fit to the raw data \cite{Riley2021, Gelman2006}. This point summarizes the debate between two-stage and one-stage individual participant data (IPD) meta-analysis \cite{Riley2021}.

Our meta-analysis treats each of the 20 datasets as an independent study. Unlike the classical setting our exchangeability assumption was built for (\Cref{Section:Background}), these datasets were not produced by independent research groups: the same pipelines and code were applied to every one by a single author. Exchangeability concerns the source of dataset-to-dataset variation in true effects, not who conducted each dataset's original recording. Our 20 datasets ranged from 8 to 109 subjects and varied in hardware and recording protocol, giving each a data-generating process distinct enough to justify treating them as separate studies.

Between-study heterogeneity for our five metrics was low by conventional thresholds, ranging from $I^2=0.45\%$ for sharpness to $28.61\%$ for reliability (\Cref{Table:Variance}). This is at least consistent with true effects coming from a shared distribution rather than each dataset behaving as its own regime. This is not proof that exchangeability holds: our largest dataset had more than ten times the subjects of our smallest, and structural differences of this magnitude could in principle produce non-exchangeable effects that a sample of 20 datasets lacks the power to detect through $I^2$ alone. We treat the low observed heterogeneity as reassuring rather than conclusive.

\subsection{Future Work}

A natural next step is a similar large-scale experiment with Bayesian partial-pooling across subjects and sessions \cite{McElreath2020, Gelman2006}. We hypothesize that partial-pooling would yield larger effects on prediction performance, with Brier score as the primary metric and reliability, resolution, discrimination, and sharpness as secondary metrics. The complete-pooling results reported here are the baseline such an experiment must beat, and their effect sizes mark the gains it must exceed. The computational costs we measured provide a reference point for the expense of fitting hierarchical models.

\section{Conclusion}\label{Section:Conclusion}

We conducted a large-scale experiment on Bayesian complete-pooling for cross-subject left/right hand classification across 20 MI-EEG datasets. It produced significant differences in reliability and sharpness, although reliability was unstable under leave-one-out analysis. Neither remained significant without the 6 datasets with flagged posterior sampling. Brier score, resolution, and discrimination showed no significant differences. In every pipeline, feature extraction was deterministic and only the classifier was Bayesian. With hyperparameters fixed rather than tuned, these results may not hold for tuned pipelines, especially neural networks. Whether partial-pooling across subjects and sessions yields larger effects is the natural next test.

\section*{Funding}
This research received no specific grant from any funding agency in the public, commercial, or not-for-profit sectors.

\section*{Conflict of Interest}
The author declares no conflict of interest.

\section*{Ethics Statement}
This study did not involve the collection of new data from human participants. All datasets analyzed were previously collected, publicly available, and gathered under the ethical approvals described in their original publications (cited in \Cref{Table:SampleCharacteristics}).

\section*{Data Availability}
The datasets analyzed in this study are publicly available through the Mother of All BCI Benchmarks (MOABB) framework. Primary sources for each dataset are cited in \Cref{Table:SampleCharacteristics}. A patch exposing predictions from MOABB, along with code used to generate posterior sampling diagnostics and energy measurements and to perform the meta-analysis reported here, is available at \url{https://doi.org/10.5281/zenodo.20415453}. Raw per-fold predictions, posterior sampling diagnostics, and energy measurements are archived at \url{https://doi.org/10.5281/zenodo.21538705}.

\bibliography{references}

\end{document}